%% file: main.tex
\documentclass[runningheads]{llncs}
\usepackage[T1]{fontenc}
\usepackage{graphicx}
\usepackage{epsfig}
\usepackage{subcaption}
\usepackage{calc}
\usepackage{multicol}
\usepackage{pslatex}
\usepackage{multirow}
\usepackage{xcolor}
\usepackage{mathtools}
\usepackage{hyperref}
\hypersetup{
    colorlinks=true,
    linkcolor=blue,
    filecolor=magenta,      
    urlcolor=cyan,
    pdftitle={VISAPP 2022 Mridula Painting},
    pdfpagemode=FullScreen,
    }

\ifx\CleanVersionFlag\undefined
    \newcommand{\HS}[1]{\textcolor{purple}{(HS: #1)}}
    \newcommand{\FL}[1]{\textcolor{blue}{(FL: #1)}}
    \newcommand{\JD}[1]{\textcolor{orange}{(JD: #1)}}
    \newcommand{\MV}[1]{\textcolor{brown}{(MV: #1)}}
\else
    \newcommand{\HS}[1]{}
    \newcommand{\FL}[1]{}
    \newcommand{\JD}[1]{}
    \newcommand{\MV}[1]{}
\fi

\begin{document}
\title{ST-SACLF: Style Transfer Informed Self-Attention Classifier for Bias-Aware Painting Classification}
\titlerunning{ST-SACLF for Bias-Aware Painting Classification}
%
\author{Mridula Vijendran\orcidID{0000-0002-4970-7723} \and
Frederick W. B. Li\orcidID{0000-0002-4283-4228} \and
Jingjing Deng\orcidID{0000-0001-9274-651X} \and
Hubert P. H. Shum $^\dag$ \orcidID{0000-0001-5651-6039}
}
\authorrunning{Vijendran et al.}
%
\institute{Department of Computer Science, Durham University\\
\email{\{mridula.vijendran, frederick.li, jingjing.deng, hubert.shum\}@durham.ac.uk}\\
}
\maketitle              
\begin{abstract}

Painting classification plays a vital role in organizing, finding, and suggesting artwork for digital and classic art galleries. Existing methods struggle with adapting knowledge from the real world to artistic images during training, leading to poor performance when dealing with different datasets. Our innovation lies in addressing these challenges through a two-step process. First, we generate more data using Style Transfer with Adaptive Instance Normalization (AdaIN), bridging the gap between diverse styles. Then, our classifier gains a boost with feature-map adaptive spatial attention modules, improving its understanding of artistic details. Moreover, we tackle the problem of imbalanced class representation by dynamically adjusting augmented samples. Through a dual-stage process involving careful hyperparameter search and model fine-tuning, we achieve an impressive 87.24\% accuracy using the ResNet-50 backbone over 40 training epochs. Our study explores quantitative analyses that compare different pretrained backbones, investigates model optimization through ablation studies, and examines how varying augmentation levels affect model performance. Complementing this, our qualitative experiments offer valuable insights into the model's decision-making process using spatial attention and its ability to differentiate between easy and challenging samples based on confidence ranking. 

\end{abstract}

\input{sections/intro-FL.tex}

\input{sections/related_works-FL.tex}

\input{sections/methodology-FL.tex}

\input{sections/experimental_results-FL.tex}

\input{sections/conclusions-FL.tex}
\bibliographystyle{splncs04}
\bibliography{main}





\end{document}

%% file: sections/intro-FL.tex
\section{Introduction}
\label{sec:introduction}






Automatic painting analysis is crucial for efficient painting indexing, retrieval, and recommendation, serving the arts industry's requirements. The transition to online art galleries due to evolving business models has spurred a demand for improved management of vast digitized artwork collections. In contrast to traditional physical galleries, online platforms struggle with the requirement to automatically analyze artworks and offer personalized recommendations to customers. This discrepancy in scalability and personalization is a central market challenge. It is against this backdrop that our focus turns to researching the intricacies of painting classification—a key component of painting analysis—as we seek to develop solutions to address these pressing issues.

\begin{figure}[!ht]
  \centering
   \includegraphics[width=0.4\linewidth]{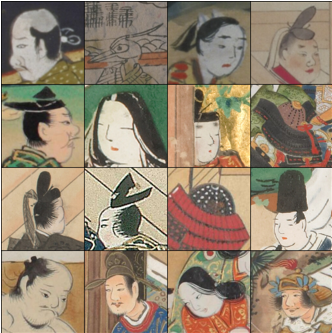}
  \caption{Image samples from the Kaokore dataset as depicted in the previous work \cite{vijendran23tackling}.}
  \label{fig:kimgs}
 \end{figure}



Deep learning methods have showcased remarkable efficacy in real-world image analysis, yet their translation to painting faces introduces a complex cross-domain generalization hurdle \cite{9847099}. Prior research endeavors have dissected various facets of this cross-domain challenge, spanning training processes, data manipulation, and model architectures. Approaches such as Transfer Learning, while effective, hinge on labeled data samples and are encumbered by the dataset size within the target domain \cite{tian2020kaokore}. 
In contrast, self-supervised learning using Contrastive Learning is more data-efficient but faces challenges when dealing with smaller, heterogeneous datasets \cite{Madhu2022}.
While using model gradient-based data augmentation is helpful for strengthening training data, it also introduces domain shifts from test or validation datasets, which could potentially cause model overfitting \cite{wang2017effectiveness,47890}. Seeking to redress data bias in the painting domain, Style Transfer-based augmentation strategies \cite{10.1007/978-3-030-84529-2_7,9577786,jackson2019style,zheng2019stada,9716108} have emerged, emphasizing style invariance and bridging domain gaps across target and source domains. However, these methodologies bypass the significance of nurturing feature-level alignment across varied abstractions, including functional and geometric insights \cite{Matthias2022}, alongside discriminative semantic-level specifics \cite{1467360,jetley2018learn}. Incorporating these aspects could greatly enhance their learning capabilities.



Data augmentation techniques have been instrumental in refining classifier training, addressing domain adaptation, and mitigating data bias in instances of class imbalance. However, these methods typically confine their influence to either the data or the model architecture. Our work seeks to harness the capabilities of style transfer in a novel way—by customizing data to harmonize with the domain learned by the model's backbone. This distinctive approach empowers us to introduce data augmentations that do not only transform the image's style and content but also align them with the model's inherent features. Consequently, this methodology serves a dual purpose: enhancing training dynamics and facilitating domain adaptation. By generating style transfer augmentations with variable proportions for both the majority and minority classes, we can strategically apply different styles to class-specific sections of the data distribution. This enables stylized minority class samples to represent the rarer instances, while the stylized majority class samples mirror the representative instances within the dataset.


In this study, we address data bias using a multi-step approach involving transforming original images into class-preserving stylized images via Adaptive Instance Normalization (AdaIN) \cite{huang2017arbitrary}. Our proposed system encompasses the classification of the model using both the original and stylized data distributions. In the initial phase, the inherent distribution within each class is conserved by applying class-specific style transfer to the images. Moreover, by customizing the quantity of augmented samples to enhance model performance, we empower the model to strategically select dataset configurations that optimize performance without disregarding vital information from the original training distribution. In the subsequent stage, the model undergoes classification using both augmented and original training data. During this phase, spatial attention is employed to identify potential data bias during clustering, thereby producing interpretable attention maps. A key innovation of our approach lies in the utilization of extracted feature maps across different levels of abstraction. This facilitates the establishment of feature correspondences spanning from global to local scales, thereby augmenting the influence of these features on the model's predictions, thereby enhancing the supervision signal. These enriched features are concatenated and directed into the classifier head. Finally, we adopt a dual-stage hyperparameter search strategy. This approach commences with a grid search, followed by a Bayesian search, and culminates in the gradual unfreezing of the resulting model. Our model optimization strategy facilitates a search that spans from the hyperparameters to the model parameter groups, transitioning from a broad exploration to a finely-tuned investigation, all accomplished with a reduced number of search trials.


We conducted extensive experiments to validate our novel model's effectiveness in addressing class imbalance, using both qualitative and quantitative analyses. The qualitative phase encompassed a detailed examination of classifier samples with varying confidence levels, scrutinizing attention map responses to assess class balancing, and gauging the influence of style and content layers. On the quantitative front, we systematically evaluated the contribution of the spatial attention layer and the data augmentation strategy. Our findings on the Kaokore dataset showcase comparable performance, where our system achieved an accuracy of 87.24\% after 40 epochs, outperforming the state-of-the-art LOOK method \cite{feng2021rethinking}, which reached 89.04\% accuracy after 90 epochs, and demonstrating its prowess with a lower parameter requirement. Fine-tuning the proportions of $p_1$ and $p_2$, we achieved a precision of 80.3\% and recall of 81.57\% using a ResNet-50 \cite{8745417} backbone. Furthermore, we scrutinized trends resulting from distinct augmentation ratios for the majority and minority classes, while evaluating classifiers of varying representation capacities to deepen our insights into their effectiveness.



Our initial findings were presented in \cite{vijendran23tackling}. This paper builds upon that foundation with a multitude of fresh enhancements. The crux of our advancements lies in the introduction of a consecutive model optimization strategy, which concurrently refines the backbone and classification network within an end-to-end framework, while optimizing the associated hyperparameters. Notably, the updated experimental results underscore a substantial improvement in network performance compared to our prior work.


Our code is available in \url{https://github.com/41enthusiast/ST-SACLF-ver1.1}. 
Our main contributions include:

\begin{itemize}
    \item We present a spatial attention classification system that achieves comparable results to the SOTA performance from the LOOK model on the Kaokore dataset. Remarkably, our approach accomplishes this with significantly reduced training time and training parameters, enhancing practical efficiency.
    
    \item To address data bias, we introduce a novel approach involving data balancing through style transfer-based data augmentation. This innovative technique draws styles from varying levels of deep features, enhancing the model's capacity to handle skewed data distributions more effectively.
    
    \item We propose a dual-stage hyperparameter search mechanism paired with fine-tuning strategies, resulting in a significant improvement of model performance. This optimization strategy enriches the system's overall capabilities by iteratively refining the model's configuration.
    
    \item Our system showcases a dynamic trade-off capability: the flexibility to adjust the augmentation ratio between rare and representative classes. This adaptability underscores the system's versatility, allowing users to tailor model performance based on specific needs.
\end{itemize}






%% file: sections/related_works-FL.tex
\section{Related Work}

Our focus is on painting classification, a field with restricted data availability. This section reviews closely related cross-domain generalization techniques, including transfer learning, data augmentation, and self-supervised learning, that are commonly used to address proposed challenge by enhancing the model, data, and training processes.


\subsection{The Model Level}
Fine-tuning pretrained models using smaller art datasets is feasible if the divergence between source and target datasets remains minimal to avoid unfavorable knowledge transfer \cite{wu2021online,zhang2022survey}. Introducing feature-level connections from different model components can blend data representations from lower to higher levels, enhancing task optimization and mitigating disturbances and defects during model refinement.

Feature-level correspondence effectively aligns semantically congruent and geometrically coherent features. Notably, the model's efficacy is boosted through the integration of the learned attention maps module \cite{jetley2018learn}. This module utilizes intermediate feature maps of the classifier to generate attention maps, which rescale local features. Subsequently, these attention maps are concatenated with the classifier's feature extractor output, facilitating varied local and global attention. Prior efforts in visual attention, such as ``show, attend, and tell'' \cite{xu2015show}, employed elementary versions of attention like soft and hard attention. Soft attention is trained using standard backpropagation but encompasses redundant black regions, while hard attention behaves akin to image cropping, being non-differentiable in nature.
%

By incorporating contextual information to capture scenes with co-occurring objects \cite{Madhu2022}, a few-shot detector employing a siamese network enhances object recognition. Through co-attention, it captures non-local features, and co-excitation forms multiple heads to capture relations across different levels of object abstraction. We employ Spatial Attention \cite{woo2018cbam} in our classifier to visualize the impact of style transfer and to retain coarse-to-fine details within images. The learned attention map is further influenced by input data amplified through selected layers. This attention mechanism serves as both a weak supervision signal \cite{jetley2018learn} and a pseudo memory bank, preserving contextual information among features fed to the module. While the spatial attention module covers a broader area due to its interaction with the spatial regions of feature maps, computational demands vary based on the chosen feature maps. In contrast, the spatial inconsistency of gradient maps makes our spatial attention more suitable for highlighting discriminative regions of interest. Multistage object detection effectively leverages discriminative features using Class Activation Maps \cite{Jeon2020}, which connect multistage models similarly to end-to-end training and offer robust system performance. However, gradient maps lack spatial consistency, rendering our spatial attention mechanism more appropriate for accentuating distinct regions of interest.

Feature engineering with template matching provide handcrafted features for comparison \cite{1467360}, and are sometimes improved with methods like voting to find the best match. Multi-style feature fusion \cite{Ufer2021}, on the other hand, employs K-means clustered style templates for image stylization. It then fuses extracted proposals to generate region descriptors, followed by dimensionality reduction using PCA to construct a search index. Iterative voting conducts local matches against this index, facilitating unsupervised object retrieval. This architecture, employing a fine-tuned VGG-16, utilizes style transfer for both proposal selection and style adaptation. It can even detect gestures \cite{Joseph2011} through correlations in position, scale, and orientation, employing non-parametric methods such as kernel density estimation. Template matching offers a feasible approach for identifying locally recurring patterns, detailed gestures inherent to artists, and other repetitive elements. However, it necessitates substantial preprocessing due to its limited coverage of variations.

Learned feature extractors like region proposals focus on extracting regions of interest as the main outcome, in contrast to methods like gradient maps or attention maps where this is a secondary result. Other multistage models, such as those detecting bounding boxes and keypoints for human figures \cite{Matthias2022}, employ semi-supervised learning with transformers and a teacher-student model. This involves distilling knowledge from the photograph domain to the realm of paintings, predicting fixed sets of proposals for each image. This eliminates the need to handle overlapping boxes and the imbalance between foreground and background. While multistage models that separate the problem into distinct transformations or formulations require a refinement stage for adding details and maintaining consistency, as well as a fusion generator to incorporate disentangled pose information, these modalities may provide geometry details that are not readily available in the painting domain due to unique rendering styles \cite{Elliot2013}.


\subsection{The Data Level}
Utilizing style transfer for data augmentation has the potential to enhance classification both at the data and feature levels \cite{organisciak20makeup}. In the past, style transfer methods were characterized by their slow and iterative optimization procedures \cite{gatys2015neural}, which altered the stylized image while keeping the model layers unchanged. However, these methods often resulted in style and content misalignment. In our case, as our model emphasizes improved painting classification through style invariance, content-specific style transfer takes a backseat in our objectives. We chose AdaIN-based style transfer due to its suitability for our needs, as more recent techniques \cite{huang2017arbitrary,Chandran_2021_CVPR,kolkin2022neural} feature separate transformation networks that enable the generation of various stylized images during inference.

At the data level, style transfer intervenes directly in the training distribution, whereas at the feature level, it modifies the model's internal representations. Techniques such as Smart Augmentation employ model features to merge samples obtained through methods like clustering \cite{7906545}, enabling the generalization of augmentation strategies between different networks. Analogously, style transfer merges images aligned with the model's features to infuse both style and content. STDA-inf \cite{10.1007/978-3-030-84529-2_7} enhances the training dataset by interpolating variations between intra-class or inter-class specific styles and the average style during training. Methods like StyleMix and StyleCutMix \cite{9577786} investigate the interplay between style and content in synthetic samples, assigning labels based on the proportion of source images. Techniques such as Style Augmentation \cite{jackson2019style} and STADA \cite{zheng2019stada} analyze the effectiveness of varying style degrees in stylized images to bolster model robustness. While STDA-inf and StyleMix bear similarities to our approach, they do not address the specific challenge of class imbalance or the task of enhancing style variation diversity by strategically selecting quality stylized images.

At the feature level in the data generation process, employing style transfer on the image generator or transformation model contributes to domain generalization and sample diversity \cite{9716108}. This technique involves introducing style as noise within the layers to generate feature maps spanning multiple source domains for the feature extractor. The classifier's training leverages both the original and augmented features. Integrating style transfer augments mitigates texture bias and fosters style invariance, aligning images with similar content but differing modalities.

Previous research has explored the synthesis of data augmentations through the manipulation of the training dataset using model gradients \cite{zhao2021DC,li2020dada,zhao2020dataset}, aiming to distill data into key model representations. Data distillation methods \cite{wang2018dataset} offer the advantage of creating a more concise and effective representation of the training data. These techniques encapsulate the training distribution into a representation optimized for the model or for shared embeddings bridging the training and target data distributions. Unlike compression-based approaches, our proposed work focuses on addressing data bias through the lens of style invariance. On the other hand, model-agnostic data augmentation methods independently or interdependently modify training data \cite{48557,49393} using existing samples. MixUp combines images with random or model-guided selections, while MixMatch applies individual traditional augmentation techniques. These methods include rotations, normalization, noise manipulation, color adjustments, and geometric transformations like shearing and translation.

\subsection{The Training Process}



Self-supervised techniques like contrastive learning \cite{islam2021broad,feng2021rethinking} leverage data similarities and differences to enhance model training efficiency, optimizing all model parameters. Contrastive learning has found applications in pre-training or CLIP embeddings \cite{conde2021clip} within the art domain. It also adapts well to address unknown instance-level deformations or noise degradation \cite{shi2021contrastive}, proving valuable for enhancing historical paintings. These methods offer supplementary supervision to refine outcomes, such as transferring and magnifying pose information from a photo to a painting \cite{Qingfu2020}. Although effective for training models on data with limited annotations, they exhibit limitations when dealing with diverse art datasets \cite{Madhu2022}. Nonetheless, these techniques have demonstrated effectiveness on our specific Kaokore dataset \cite{islam2021broad}.

Our approach also sets itself apart from competitors who utilize contrastive learning \cite{islam2021broad,feng2021rethinking} to leverage data similarities and differences for improved model training. In contrast, our classifier backbone relies on pretrained models \cite{canziani2016analysis,8745417} initially trained on another task, with their heads subsequently fine-tuned for painting classification. These models \cite{Pramook2016} can be further adapted to smaller drawing datasets to effectively address the domain gap issue often present in pretrained models trained on photographic data.
Certain iterations of MCNC \cite{Sizyakin2020}, a model employing image processing morphological operations, demonstrate generalization to new data through fine-tuning within a two-stage model, yielding more precise boundary detection. However, fine-tuning on significantly larger backbones compared to the model's head \cite{Jeon2020} can lead to inefficiencies due to the freezing of substantial portions of the network, often resulting in suboptimal outcomes. In contrast, our approach introduces increments to distinct stages of the model without causing interference between them. Alternatively, distilling geometric information for domain adaptation yields improved outcomes \cite{Matthias2022} compared to fine-tuning or style transfer accompanied by additional label conditioning. By employing data augmentation, we mitigate negative transfer effects by narrowing the domain gap between the source and target datasets.

The selection of augmentations can also be adapted to leverage the model's inherent tendencies \cite{47890,wang2017effectiveness}. For instance, a GAN-based style transformation network \cite{wang2017effectiveness} employs meta learning to acquire augmentations within a small network that can subsequently generalize to a larger network. Exploration into diffusion models has also been pursued to represent style features \cite{zhang23diffusion}. Autoaugment \cite{47890}, in contrast, employs reinforcement learning policies to choose augmentations. These policy-based augmentations are drawn from a selection pool encompassing traditional image augmentations. The chosen augmentations reflect domain-level insights and introduce biases aligned with the model's architecture. In our system, we intervene at the data level, randomly selecting styles from the same class to uphold intraclass distribution and counteract sampling bias by introducing varying amounts of data to each class.


%% file: sections/methodology-FL.tex
\section{Methodology}

Our innovative solution overcomes limitations in existing augmentation techniques. Unlike prior methods, it tackles class imbalance in interclass scenarios, allowing flexibility to prioritize performance or bias reduction. Our style transfer-based augmentation fine-tunes both style and content for task alignment, offering an effective approach for classification and bias challenges. Our proposed system addresses these issues through the following features:
\begin{itemize}
    \item \textbf{Balancing Bias and Enhancing Model}: By incorporating varied amounts of style-enhanced data into both majority and minority classes, we mitigate bias and boost model performance. These augmentations also cultivate texture consistency across diverse styles in each sample, compelling the model to prioritize image content.
    
    \item \textbf{Aligning Style and Precision}: We ensure that the level of detail from style transfer aligns with our model's classification needs. Achieved through spatial attention modules, this alignment emphasizes local features, minimizing performance differences caused by varying style transfer configurations during classification.
\end{itemize}

\begin{figure}[!ht]
  \centering
   \includegraphics[width=\linewidth]{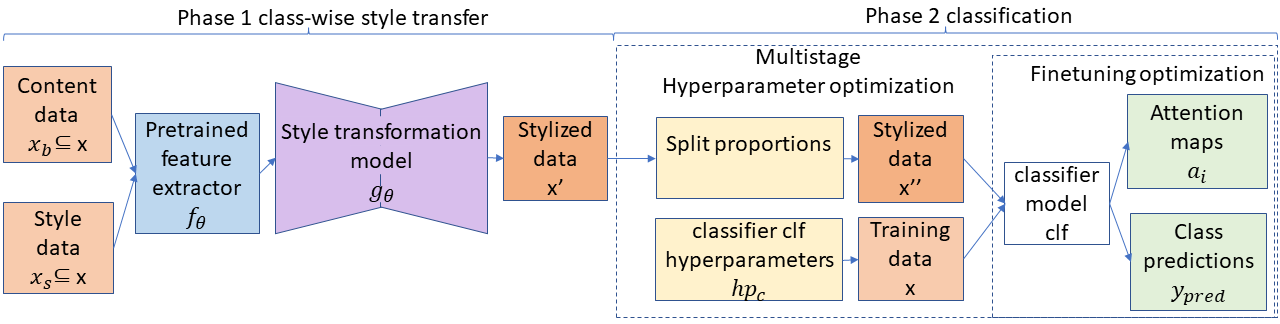}
    \caption{Our system for style-based data augmentation to enhance model classification.}
  \label{fig:overall}
 \end{figure}

The system is sequentially divided into two phases, as shown in Figure \ref{fig:overall}. Phase 1 involves the core process of style transfer, which transforms the original training data into augmented versions. This transformation entails the uniform selection of random content-style image pairs from each class, resulting in hybrid samples. Moving to Phase 2, the augmented and original datasets merge to become the input for the classifier. Comprising a pre-trained network and a head fine-tuned with combined local and global spatial attention modules, the classifier's performance is optimized. This optimization occurs via a meticulous two-stage procedure. Firstly, grid searches are executed, meticulously exploring split proportions for stylized dataset stratification and classifier hyperparameters. Subsequently, the second stage employs Bayesian search to further refine the model. Notably, the style transfer process maintains consistency with the VGG-19 backbone, while the classifier maintains the flexibility to integrate a range of pre-trained backbones for enhanced adaptability and customization.

\subsection{Data Augmentation from Style Transfer}
To enhance the objectivity of style image selection, we present an automated approach that sets itself apart from STaDA \cite{zheng2019stada}. This new method addresses the challenge of subjective style image choices.
\begin{figure*}[!ht]
  \centering
   \includegraphics[width=\textwidth]{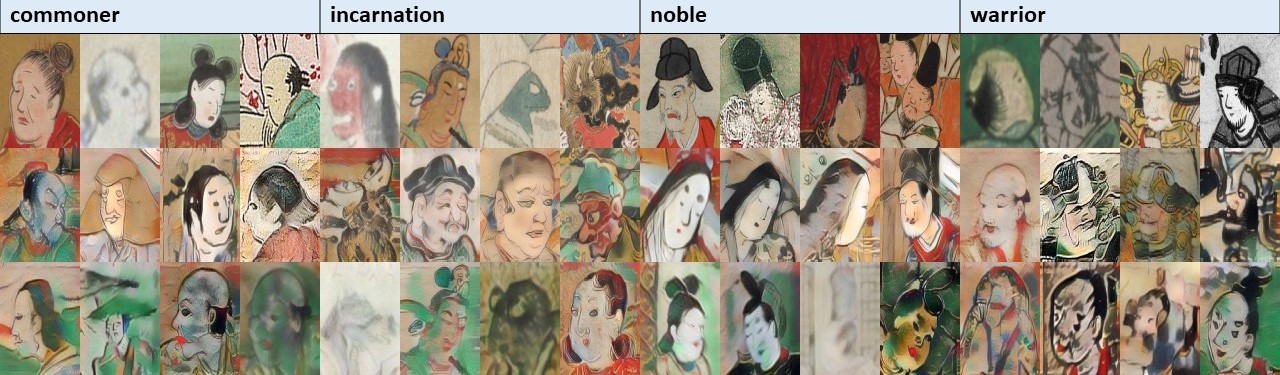}
  \caption{The original samples per class followed by good and sub-optimal style transfer augmentations in the second and third rows, respectively. The augmentations and original images are borrowed from the previous work \cite{vijendran23tackling}.}
  \label{fig:augs}
 \end{figure*}

For our data augmentation strategy, we propose the utilization of Adaptive Instance Normalization (AdaIN) \cite{huang2017arbitrary} within an image transformation network, which ensures rapid transformation speed. AdaIN operates by aligning the mean and covariance of the content feature map to those of the style feature map, effectively merging information from both inputs. However, it's important to note that the transferred textures from the style image may not seamlessly align with the content image, as this process lacks context awareness. Additionally, the configuration of the transformation network is specific to the resultant textures and relies on a specially trained VGG-19 backbone.

\begin{figure*}[!ht]
  \centering
   \includegraphics[width=\linewidth]{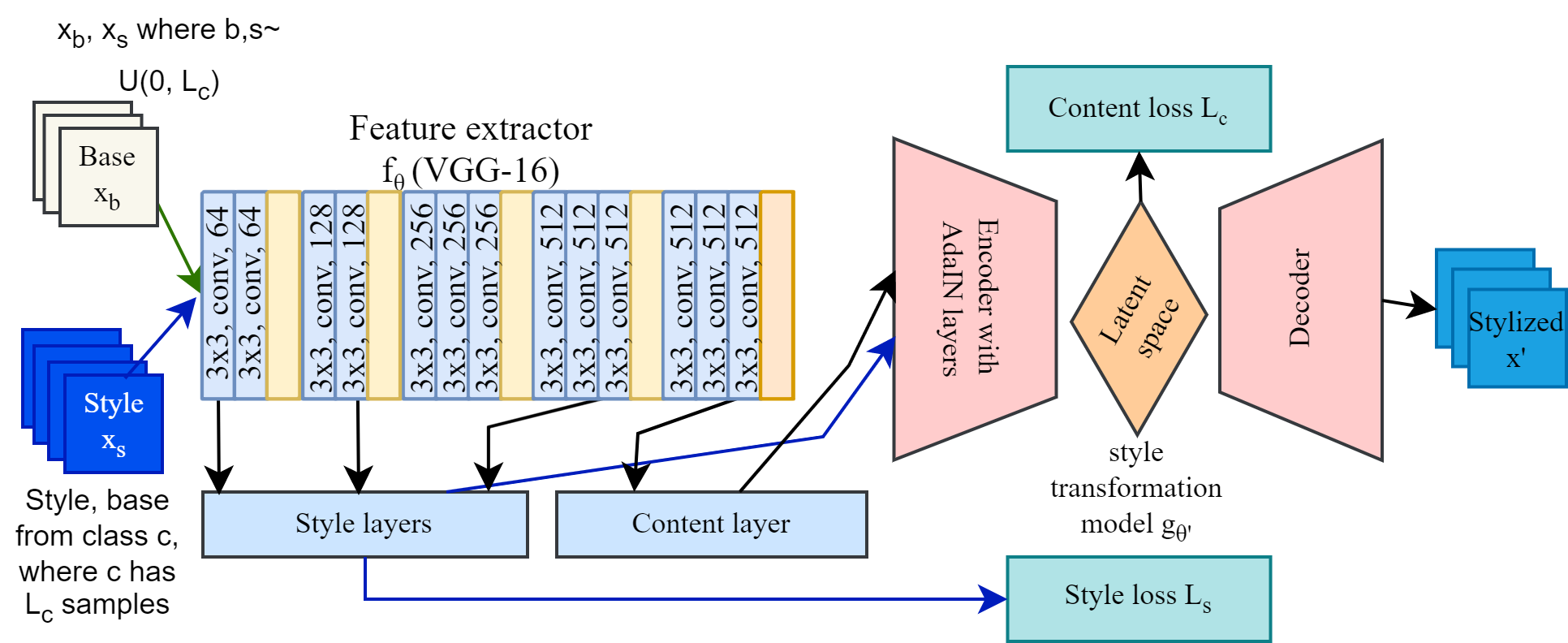}
  \caption{The style transfer model generates stylized versions of the input data per class.}
  \label{fig:st}
 \end{figure*}

By incorporating style transfer, we address domain disparities between the original training dataset containing real-world images and paintings. These discrepancies encompass variations in low-level attributes like textures, patterns, and strokes, as well as higher-level characteristics such as distinct shapes. Introducing style invariance enables the reduction of the significant domain gap that often poses challenges in fine-tuning and data generalization \cite{yosinski2014transferable}. Leveraging style transfer allows us to obscure the dataset's style, artistic semantics \cite{10.1109/TMM.2020.3009484}, and distortions, thus bridging the domain gap during transfer learning. This integration compels the classifier to rely on shared content information present in both the source and target datasets, given that convolutional neural networks exhibit heightened sensitivity to texture information \cite{von2021self}. Through data augmentations, content information specific to real and abstracted representations is disentangled, enabling superior utilization of higher-level features for classification \cite{geirhos2018imagenettrained}. This approach of style transfer and data augmentation effectively aligns domain characteristics and optimizes feature deployment, enhancing overall model performance.

We introduce data-augmented versions of the training data, generated prior to training using the same model, an alternative to Smart Augmentation \cite{7906545}. To counter class imbalance through style transfer, we adopt a straightforward heuristic wherein half of the classes with the highest sample counts are labeled as representative ($p_1$), while the remaining classes are deemed rare ($p_2$). This designation governs the proportion of augmented samples incorporated into the representative and rare classes. Importantly, our data augmentation approach does not demand an encoder like GANs to emphasize specific feature-level details, nor does it necessitate a separate network for training augmentation strategies aimed at the primary classification network. This methodology efficiently addresses class imbalance and data augmentation concerns while maintaining simplicity and effectiveness.

The style transfer model, as illustrated in Figure \ref{fig:st}, optimizes the style loss by leveraging the gram matrix of its feature embeddings to capture second-order statistics related to texture and feature variance. At the bottleneck of the image transformation model, the content loss is calculated, incorporating style modulation at the Adaptive Instance Normalization \cite{huang2017arbitrary} layers along with content from the reconstruction loss to train the decoder end of the transformation model. The encoder relies on a customized pre-trained VGG-19 model with normalized weights. For training the style transfer model, we utilize uniformly sampled style data from the entire dataset, exposing the model to diverse style variations. Following the training of the decoder on domain-specific training images, the style transfer can be swiftly computed during inference by employing uniformly sampled content and style images with repetitions per class. This decoder facilitates matching style statistics and image content in the feature space. Through these aligned features, we achieve versatile stylization using the AdaIN operation (Equation \ref{eq1}), which modulates the content feature using the style statistics at the output of the style transformation network's encoder. This approach allows us to efficiently incorporate style transfer as an integral part of our data augmentation strategy.
\begin{equation}
\begin{aligned}\label{eq1}
    &c = f_c(x_b)\\
    &s = f_s(x_s)\\
    &AdaIN(c,s)=\sigma(s)\left(\frac{c-\mu(c)}{\sigma(c)}\right)+\mu(s)\\
    &t = AdaIN(c,s)
\end{aligned}
\end{equation}
where c and s are content and style features from the feature extractor, respectively. $\sigma$ is the variance and $\mu$ is the mean, respectively. $t$ is the AdaIN output. $x_b$ is the content image, $x_s$ is the style image, $f_s$ and $f_c$ are the feature extractor of the pretrained model until the desired style or content layer respectively.

The content loss $L_c$ and the style loss $L_s$ for for style transfer are given as MSE losses and are computed as follows:

\begin{equation}
    \begin{aligned}
    \label{contentandstyleloss}
    &L_c = ||f(g(t) - t)||_2\\
    &L_s = || \mu(\phi_i(g(t))) - \mu(\phi_i(x_s))||_2 +
            \sum_{i=1}^L || \sigma(\phi_i(g(t))) - \sigma(\phi_i(x_s))||_2
    \end{aligned}
\end{equation}
where $t$ is the AdaIN output from Equation \ref{eq1} and content target, $x_s$ is the style image, $f$ is the encoder, $g$ is the decoder, $\phi_i$ are the style layers. The style loss matches the mean and standard statistics between the style image and the stylized image. The content loss matches the stylized features to the target features.

In the process of style transfer, our approach focuses on updating only the decoder weights during training. By extracting style and content features from their respective chosen layers, we form a stylized tensor using the AdaIN layer. The level of detail retained in the stylized tensor, be it style or structural information, is regulated by an alpha value. This tensor then undergoes decoding to generate a hybrid image, which simultaneously maintains its structural integrity and adopts stylistic attributes. This is achieved by employing content loss to ensure the hybrid image matches the content embedding and style loss to align its embeddings with the style image's characteristics. The interplay of these content and style losses guides the learning process of the decoder, ultimately influencing the formation of the hybrid image.

The quality of the generated samples per class is illustrated in Figure \ref{fig:augs}. Since most images are face-centered, the style transfer effectively transfers textures while preserving the underlying content. Nonetheless, due to the lack of specific constraints on the content transfer, certain color bleeding is observed in the stylized images, as depicted in the bottom row. In the case of the Kaokore dataset, characterized by a prevalence of green backgrounds and characters with green clothing, this color tends to bleed into the generated samples, reflecting the inherent characteristics of the dataset.

\subsection{Spatial Attention Based Image Classifier}

\begin{figure}[!ht]
  \centering
   \includegraphics[width=\textwidth]{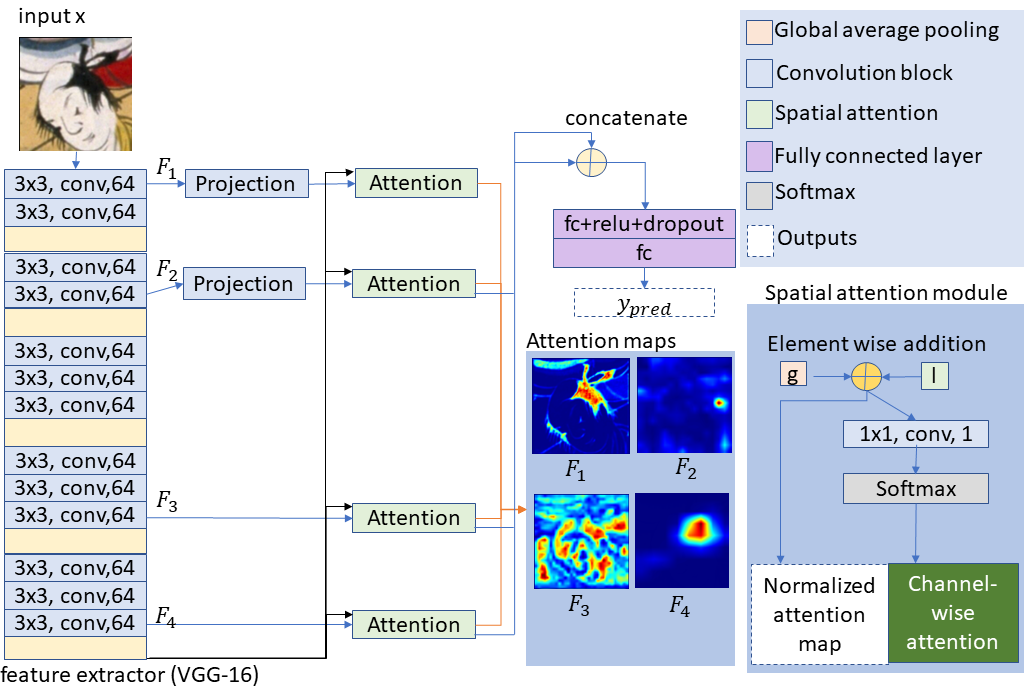}
  \caption{The classifier architecture is depicted with the model flow from the input to the outputs. The blue line indicates local features while the red line indicates global features. The output from the spatial attention layer to the fully connected layers are global features weighted by the corresponding local features.}
  \label{fig:classifier}
 \end{figure}

The classifier, as depicted in Figure \ref{fig:classifier}, is constructed using a pre-trained image classification model, such as VGG-16 and ResNet-50 \cite{canziani2016analysis,8745417}. In the VGG variants, we carefully select three layers between the initial and final layers, capturing spatially enriched features that contribute to a balanced fusion of style and content information for the classification loss. For ResNet variants, we incorporate the initial conv layer and outputs from different stages with basic and bottleneck blocks as local layers. This classifier encompasses a pre-trained backbone, projection layers, spatial attention modules, and a fully connected head. The projection layers harmonize channel features from the backbone's response maps to serve as input for the spatial attention modules. These modules compute attention based on the re-projected layer and global bottleneck features, promoting comprehensive information integration. The resultant concatenated features then pass through the head, comprising dense layers and dropout, facilitating image classification. Notably, our approach omits batch normalization layers and global training statistics that are typically leveraged in previous work for domain adaptation \cite{frankle2020training}. Instead, our approach utilizes data augmentation for domain adaptation, tackling the class imbalance challenges of the Kaokore dataset and offering a model-agnostic strategy for domain adaptation and data bias mitigation.


The spatial attention module calculates attention maps by considering both the local response map and the global feature extracted from the end of the feature extractor. This integration ensures that both local and global contexts of the image are embedded in the attention mechanism. In our approach, style transfer layers take precedence in the loss computation, leveraging the concatenated spatial attention responses at the fully connected head. Unlike activation or gradient-based attention maps, spatial attention effectively highlights important regions while suppressing background noise \cite{jetley2018learn}. This attribute does not only enhance model performance but also enable compatibility with non-attention-based classifiers. Also, spatial attention retains contextual relationships even after style transfer \cite{chang2021exploiting}, safeguarding against non-context-preserving style transfer augmentations from negatively impacting model performance. The adaptable nature of spatial attention heads accommodates varying feature map quantities in the extracted set, supporting different dimensional requirements. Our classifier head incorporates a projection layer and spatial attention module for each feature map extracted from the backbone. This design ensures compatibility with diverse feature channel sizes in the extracted feature map set, distinguishing our approach from the original work \cite{jetley2018learn}.

The utilization of focal loss is a widespread strategy in object detection tasks aimed at addressing the skewed distribution of positive and negative samples \cite{arkin2021survey} and enhancing model accuracy. This loss function effectively reduces the emphasis on easy examples during training, facilitating smoother convergence across multiple epochs with a specific hyperparameter configuration. As such, we employ focal loss as the classification loss in our spatial attention classifier. This choice helps counteract the class imbalance present in the Kaokore dataset. The formulation of focal loss is as follows:

\begin{equation}\label{focal_loss}
    \begin{aligned}
    &p_t = softmax(y_{pred})\\
    &softmax(y_{pred}) = \frac{\exp^{y_{pred}}}{\sum_{j=1}^{c}\exp^{{y_{pred}}_j}}\\
    &FL(p_t) = -\alpha(1-p_t)^\gamma y\log(p_t)\\
    \end{aligned}
\end{equation}

\subsubsection{Classification Backbone Models}
In our classification backbone models, we opt for VGG and ResNet variants as pretrained backbones for the classifiers, while employing a modified-weight VGG-19 for style transfer. The VGG variants allow us to explore the significance of style \cite{wang2021rethinking} relative to feature embeddings \cite{tian2020rethinking}, in contrast to ResNet variants. When it comes to style transfer, we utilize a set of extracted feature maps, primarily optimizing gram matrices in the style loss. This approach leads to varying stylizations based on the chosen layers. As a result, the residual connections in ResNet, which produce layers with prominent peaks and lower entropy and spread information across multiple layers or channels, are less suited for effective style transfer. We observe a significant degradation in stylization performance when substituting VGG with models that excel in the classification task, like ResNet \cite{wang2021rethinking}. Given that larger-capacity models within the same architecture type tend to correlate with improved classifier performance \cite{he2016deep} in the case of ResNet series, we choose them as the backbones for our paintings classification task. The embeddings of the ResNet models make them effective feature extractors, particularly suitable for fine-tuning on small datasets when combined with a straightforward linear head \cite{tian2020rethinking}.

\subsection{Successive Model Optimization}

A substantial enhancement in the classifier's performance is achieved by combining hyperparameter search and fine-tuning. The initial step involves utilizing Grid search to thoroughly explore various hyperparameter scales. This process effectively narrows down the scope, identifying the most optimal hyperparameter combination for the task and creating an initial foundation for focused exploration. Since model hyperparameters operate independently, we employ TPE-based Bayesian optimization \cite{bergstra2011algorithms} to suggest suitable ranges for further exploration after each trial. To streamline computations, this search is conducted on the classifier with a fixed backbone, concentrating on the parameter range relevant to learned higher-level features. This approach does not only facilitate optimization but also conserve computational resources by executing both optimization methods with a frozen backbone. Ultimately, the two-stage hyperparameter search method exhaustively explores fewer parameter combinations within the model-space, leading to enhanced performance outcomes. 

In addition to the hyperparameter optimization, we refine the model weights for the task through a gradual unfreezing of gradients alongside decreasing learning rates, progressively unfreezing higher layers first while retaining lower-level information learned by the model \cite{peters2019tune}. Employing lower learning rates is pivotal to maintaining the model's grasp on foundational knowledge during fine-tuning, thereby aiding its adaptation to the new domain. The first fine-tuning stage involves freezing the backbone and selecting the model with the best performance on the unaugmented validation set, ensuring alignment with the original training data distribution. In the second stage, the entire model is unfrozen, and the learning rate is scaled down by a factor of 10. This process entails training the model anew using the weights obtained from the first stage. Although this two-stage fine-tuning approach requires twice the effective training epochs, it yields comparable results to state-of-the-art approaches \cite{islam2021broad}.

%% file: sections/experimental_results-FL.tex
\begin{figure}[!ht]
  \centering
   \includegraphics[width=.95\linewidth]{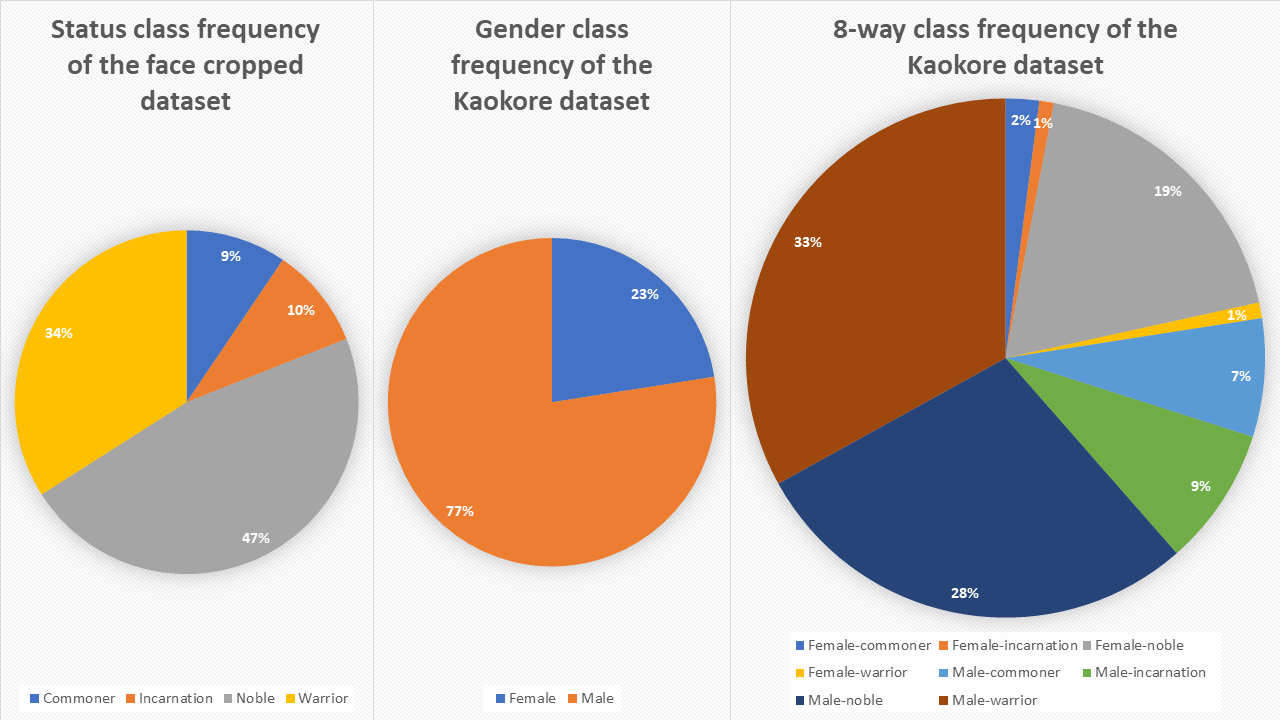}
  \caption{Class imbalance in the Kaokore dataset for status 
 , gender and an 8-way combination of both their subclasses as a classification task.}
  \label{fig:kaokore-ds}
 \end{figure}

\begin{figure}[!ht]

 \begin{subfigure}[b]{\textwidth}
  \centering
  \includegraphics[width=.7\linewidth]{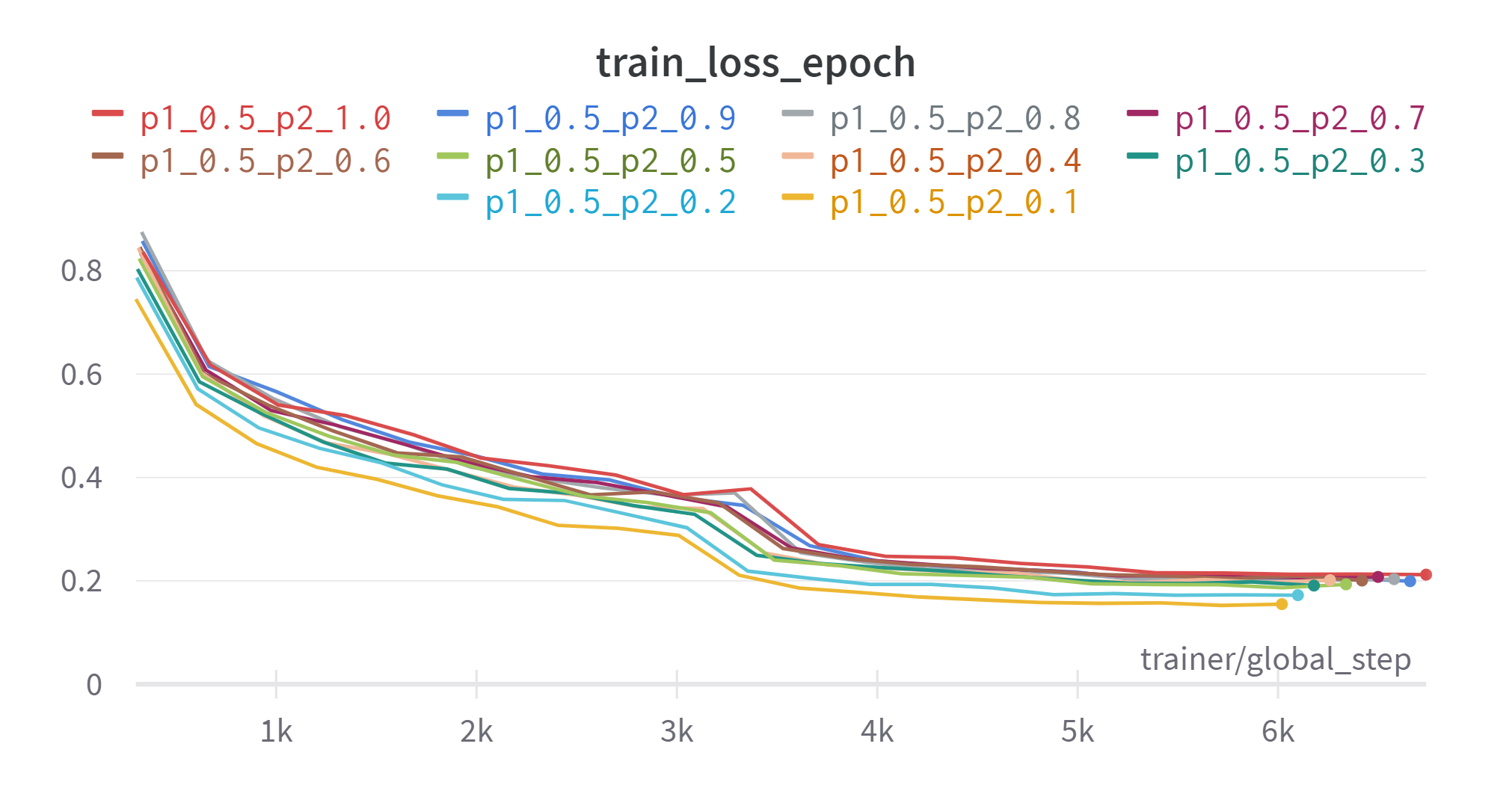}
  
  
 \end{subfigure}

 \begin{subfigure}[b]{\textwidth}
  \centering
   \includegraphics[width=\linewidth]{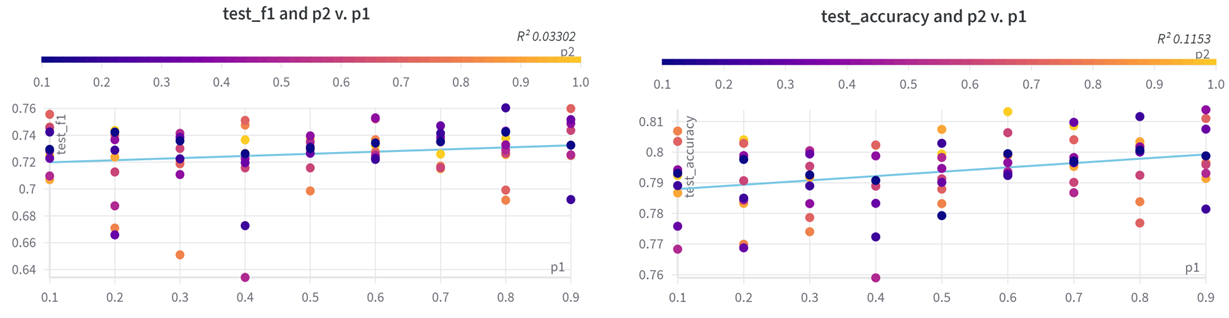}
  \end{subfigure}
 \caption{Plots depicting training convergence under varying proportions of rare samples ($p_2$) concerning train loss per epoch, test accuracy, and F1 scores. These results are compared against different levels of rare ($p_1$) and representative ($p_2$) augmentations, as discussed in \cite{vijendran23tackling}. The depicted trends showcase model performance changes with different degrees of style transfer augmentation. 
 Here, $p_1$ and $p_2$ denote the percentages of additional training data in the common classes (noble and warrior) and the rare classes (incarnation and commoner), respectively.
 The division assigns half of the classes to the common classes with the least data and the other half to the rare classes.}

 \label{fig:trends}
\end{figure}
 
\section{Experiments}

In this section, we conduct a comprehensive evaluation by comparing various data augmentation strategies and classifier optimization methods. We introduce dataset variations used in our experiments. We also perform qualitative and quantitative experiments to visually analyze the components of our system and conduct a thorough comparative study.  Finally, we provide our implementation details.

Section \ref{subsec:datasets} presents the Kaokore dataset and its variants, which serve as the foundation for our experiments in Sections \ref{subsec:qualres} and \ref{subsec:ablationstudies}. The quantitative analyses encompass comparison and ablation studies (Section \ref{subsec:ablationstudies}) to assess the efficacy of our system's modules, stylization combinations, and optimization strategies. Complementing this, the qualitative investigations (Section \ref{subsec:qualres}) delve into model interpretability via examination of the classifier's selected attention layers and confidence scores. Ultimately, Section \ref{subsec:implementationdets} outlines the system's configuration details.
 
\subsection{Datasets}\label{subsec:datasets}

The Kaokore dataset \cite{tian2020kaokore} consists of a collection of Japanese paintings, primarily centered around faces, and categorized based on gender and status attributes. This dataset offers a diverse array of facial features, encompassing varying shapes, poses, and colors, making it a suitable choice for enhancing classification performance by fostering style invariance. The gender category is further divided into male and female subclasses, while the status category is subdivided into commoner, noble, incarnation (non-human or avatar), and warrior. As depicted in Figure \ref{fig:kaokore-ds}, the dataset suffers from significant class imbalance. It predominantly comprises cropped face images, as shown in Figure \ref{fig:kimgs}. In our experiments, we emphasize the status categorization to better showcase the impact of style transfer on classification. This choice is motivated by the fact that it necessitates more finesse than hyperparameter tuning or model regularization techniques, as is the case with the gender classification task. Despite its relatively modest size, with 6,756 training images and 845 validation and test images of the same size, the dataset stands to benefit from transfer learning.

Our experiments involve two versions of the Kaokore dataset. The first version comprises the status classes \cite{tian2020kaokore}, while the second version encompasses an 8-way combination of status and gender classes \cite{islam2021broad}. This extended version introduces more class imbalance due to the presence of subclasses that are shared between the primary status and gender categories.



\subsection{Quantitative Results}
\label{subsec:ablationstudies}

We conducted a series of experiments to evaluate the effectiveness of style transfer as a data augmentation technique. Firstly, we compared our method against state-of-the-art approaches. Next, we investigated the model's performance across different data augmentation configurations of $p_1$ and $p_2$. Subsequently, we analyzed the impact of style transfer augmentation across various model capacities and architectures. Finally, we delved into experimenting with model optimization techniques to enhance the performance of the pretrained backbone, which is common to all our competing methods.


\begin{table}[ht]
\centering
\footnotesize
\caption{A comparative analysis of the Kaokore dataset's status classification using exclusively the stylized datasets with optimized hyperparameters and varying backbones, as explored in prior work \cite{vijendran23tackling}. Our data augmentation technique can seamlessly enhance the performance of all existing state-of-the-art methods. Our approach involves a two-stage hyperparameter search and the freezing of pretrained backbones.}
\label{table3}
\begin{tabular}{|l|l|l|l}
\cline{1-3}
Method                                                  & \begin{tabular}[c]{@{}l@{}}Test accuracy\end{tabular} & \begin{tabular}[c]{@{}l@{}}Trainable parameters (in millions)\end{tabular}                     &  \\ \cline{1-3}
\begin{tabular}[c]{@{}l@{}}VGG-11 \cite{tian2020kaokore}\end{tabular}    & 78.74\%                                                  & 9.2 M                                                                                          &  \\ \cline{1-3}
\begin{tabular}[c]{@{}l@{}}AlexNet \cite{tian2020kaokore}\end{tabular}    & 78.93\%                                                  & 62.3 M   
&  \\ \cline{1-3}
\begin{tabular}[c]{@{}l@{}}DenseNet-121 \cite{tian2020kaokore}\end{tabular}    & 79.70\%                                                  & 7.6 M       
&  \\ \cline{1-3}
\begin{tabular}[c]{@{}l@{}}Inception-v3  \cite{tian2020kaokore}\end{tabular}    & 84.25\%                                                  & 24 M     
&  \\ \cline{1-3}
\begin{tabular}[c]{@{}l@{}}ResNet-18 \cite{tian2020kaokore}\end{tabular}    & 82.16\%                                                  & 11 M            &  \\ \cline{1-3}
\begin{tabular}[c]{@{}l@{}}MobileNet-v2  \cite{tian2020kaokore}\end{tabular}    & 82.35\%                                   & 3.2 M                                               &  \\ \cline{1-3}
\begin{tabular}[c]{@{}l@{}}ResNet-34 \cite{tian2020kaokore}\end{tabular} & 84.82\%                                                 & 21.3 M                                                                                                                &  \\ \cline{1-3}
SelfSupCon \cite{islam2021broad}                                                                       & 88.92\%                                                  & 47 M                                                                                          &  \\ \cline{1-3}
CE+SelfSupCon \cite{islam2021broad}                                                                     & 88.25\%                                                  & 27.9 M                                                                                        &  \\ \cline{1-3}

LOOK (ResNet-50) \cite{feng2021rethinking}                                                                          & \textbf{89.04} \%                                                   & 23.5 M                                                                                                                     &  \\ \cline{1-3}

\textbf{Ours (VGG-16 backbone) }                                                                    & 82.06\%                                                 & \textbf{1.2 M }                                                                                                                &  \\ \cline{1-3}
\textbf{Ours (ResNet-34 backbone)  }                                                           & 81.38\%                                                 & \textbf{1.2 M}                                                                                                                 &  \\ \cline{1-3}
\textbf{Ours (ResNet-50 backbone)  }                                                           & 83.22\%                                                 & 20.1 M                                                                                                                 &  \\ \cline{1-3}

\end{tabular}
\end{table}

\begin{table}[ht]
\centering
\scriptsize
\caption{Performance metrics (accuracy/precision/recall) of the ResNet-34 model for various configurations of $p1$ and $p2$, each incrementing by 10\%. In this setup, the model's pretrained backbone remains frozen, with no further model fine-tuning conducted; only hyperparameter search is applied. Here, $p1$ signifies the percentage of additional data from the majority class, while $p2$ denotes the percentage of supplementary data from the minority class.}

\begin{tabular}{|l|l|l|l|l|l|}
\hline
$p1$/$p2$ & 0.1               & 0.2               & 0.3               & 0.4               & 0.5        \\ \hline
0.1   & 79.31/76.89/70.55 & 78.91/75.46/73.23 & 77.58/76.33/69.54 & 79.42/74.90/71.50   & 76.83/73.90/69.43\\\hline
0.2   & 79.76/75.93/72.82 & 78.50/75.36/71.26  & 76.88/72.01/64.97 & 79.89/74.56/72.92 & 78.44/74.16/65.81\\\hline
0.3   & 79.26/74.29/72.97 & 78.90/73.75/71.04  & 79.94/74.54/73.24 & 78.32/73.97/69.23 & 80.05/74.75/73.61\\\hline
0.4   & 79.08/75.22/70.69 & 77.23/74.38/65.60  & 78.33/73.23/70.98 & 79.88/75.09/70.38 & 75.90/72.43/61.49\\\hline
0.5   & 77.93/74.17/72.21& 80.29/76.20/71.16 & 79.02/73.37/72.14& 79.47/76.59/71.95& 79.83/75.52/72.04\\\hline
0.6   & 79.95/76.04/71.71 & 79.24/73.31/71.28 & 79.30/74.03/70.97  & 79.66/76.91/73.98 & 79.36/73.41/71.73\\\hline
0.7   & 79.71/74.61/72.64 & 79.66/76.40/72.43  & 80.97/76.59/73.26 & 79.83/74.28/73.45 & 78.68/75.03/72.38\\\hline
0.8 & 80.07/76.38/72.70  & 81.16/77.31/\textbf{75.03} & 80.00/75.29/73.46    & 80.17/74.82/72.12                          & 80.05/76.12/70.87\\\hline
0.9 & 79.88/75.88/71.76 & 78.14/72.34/69.05                           & 80.75/77.13/73.84 & \textbf{81.38}/76.40/73.83 & 79.31/74.65/71.06\\\hline
1.0& 77.87/73.32/66.63& 79.78/76.16/73.50 & 78.15/73.47/70.27& 79.71/74.37/70.57& 79.36/74.72/70.42\\\hline

\hline
\hline
$p1$/$p2$ & 0.6               & 0.7               & 0.8               & 0.9               & 1.0               \\ \hline
0.1   & 79.37/76.03/73.34 & 80.35/75.96/75.21 & 80.69/76.44/73.23 & 78.67/73.10/69.13  & 79.24/73.61/71.61 \\ \hline
0.2   & 79.07/72.84/70.49 & 80.29/75.77/72.66 & 76.99/70.92/65.76 & 78.33/74.40/70.89  & 80.40/75.47/73.41  \\ \hline
0.3   & 79.54/74.75/73.61 & 79.54/74.64/71.91 & 77.87/73.35/70.67 & 77.40/70.26/64.99  & 79.19/73.93/71.10  \\ \hline
0.4   & 78.89/72.41/71.15 & 80.24/76.94/73.60  & 80.23/76.05/73.82 & 78.91/73.98/71.69 & 80.23/75.69/72.14 \\ \hline
0.5   & 79.13/73.64/70.11 & 78.79/73.61/72.40  & 78.32/72.86/68.59 & 80.74/76.19/71.37 & 79.94/74.82/72.98 \\ \hline
0.6   & 80.64/77.58/73.53 & 79.36/76.22/70.69 & 79.90/75.16/72.69  & 79.25/73.70/71.91  & 81.32/76.04/71.52 \\ \hline
0.7   & 79.02/75.24/69.63 & 80.40/75.64/72.17  & 78.68/73.88/70.29 & 79.53/73.71/70.04 & 80.86/75.89/71.19 \\ \hline
0.8   & 79.25/75.32/70.70  & 77.69/71.70/68.91  & 78.38/73.22/66.91 & 80.34/74.29/71.60  & 80.23/75.30/72.74  \\ \hline
0.9   & 79.59/75.63/73.29 & 81.10/77.70/74.61   & 79.66/73.98/71.43 & 79.14/74.34/70.97 & 79.08/74.85/71.52 \\ \hline
1.0 & 78.15/77.43/62.81 & 80.41/\textbf{78.62}/73.60 & 80.24/76.39/71.44 & 80.29/76.37/73.45 & 78.96/74.33/69.42 \\ \hline
\end{tabular}
\label{prob_sweep}
\end{table}

\begin{table}[ht]
\centering
\caption{Test performance for different model capacities for our ResNet backbone variations for the 8-way Kaokore dataset version. We do not perform the 2-stage model optimizations and use the best configuration from the ResNet-50 variation. }\label{tab:table5}
\begin{tabular}{|l|l|}
\hline
ResNet backbone & Test accuracy \\ \hline
Ours (ResNet-34 backbone)      & 79.87         \\ \hline
Ours (ResNet-50 backbone)      & 80.94         \\ \hline
Ours (ResNet-101 backbone)     & \textbf{80.95}         \\ \hline
Ours (ResNet-152 backbone)     & 80.47         \\ \hline
\end{tabular}
\end{table}

\begin{table}[ht]
\centering
\footnotesize
\caption{Comparison of model performance using various classifier backbones with and without data augmentation. The best mixes are determined through a grid search involving multiple rare and representative proportions, applied to a classifier with a frozen backbone. This analysis isolates the impact of data augmentation alone, without employing the two-stage model optimization process.}
\begin{tabular}{|l|l|llll|}
\hline
\multirow{2}{*}{\begin{tabular}[c]{@{}l@{}}Model Architecture \end{tabular}} &
  \multirow{2}{*}{\begin{tabular}[c]{@{}l@{}}Data augmentation setup\end{tabular}} &
  \multicolumn{4}{l|}{Metrics (in percentage)} \\ \cline{3-6} 
                          &                                  & \multicolumn{1}{l|}{Accuracy} & \multicolumn{1}{l|}{Recall} & \multicolumn{1}{l|}{Precision} & F1 score \\ \hline
\multirow{2}{*}{VGG16}    & \begin{tabular}[c]{@{}l@{}}No augmentation\end{tabular} & \multicolumn{1}{l|}{79.91}         & \multicolumn{1}{l|}{71.08}       & \multicolumn{1}{l|}{73.09}          & 72.00    \\ \cline{2-6} 
                          & Optimal rare and representative mix                          & \multicolumn{1}{l|}{82.06}    & \multicolumn{1}{l|}{71.41}  & \multicolumn{1}{l|}{75.90}     & 73.27         \\ \hline
\multirow{2}{*}{VGG19}    &\begin{tabular}[c]{@{}l@{}}No augmentation\end{tabular} & \multicolumn{1}{l|}{78.84}         & \multicolumn{1}{l|}{67.67}       & \multicolumn{1}{l|}{73.34}          & 69.80 \\ \cline{2-6} 
                          & Optimal rare and representative mix                          & \multicolumn{1}{l|}{80.68}    & \multicolumn{1}{l|}{71.43}  & \multicolumn{1}{l|}{74.06}     & 72.39            \\ \hline
\multirow{2}{*}{ResNet34} & \begin{tabular}[c]{@{}l@{}}No augmentation\end{tabular} & \multicolumn{1}{l|}{80.03}         & \multicolumn{1}{l|}{71.49}       & \multicolumn{1}{l|}{75.93}          & 73.29  \\ \cline{2-6} 
                          & Optimal rare and representative mix                          & \multicolumn{1}{l|}{81.38}    & \multicolumn{1}{l|}{73.83}   & \multicolumn{1}{l|}{76.40}     & 74.88           \\ \hline
\multirow{2}{*}{ResNet50} & \begin{tabular}[c]{@{}l@{}}No augmentation\end{tabular} & \multicolumn{1}{l|}{78.43}         & \multicolumn{1}{l|}{69.55}       & \multicolumn{1}{l|}{71.58}          & 70.48    \\ \cline{2-6} 
                          & Optimal rare and representative mix                          & \multicolumn{1}{l|}{\textbf{87.24}}    & \multicolumn{1}{l|}{\textbf{81.57}}   & \multicolumn{1}{l|}{\textbf{76.90}}     & \textbf{79.97}         \\ \hline
\end{tabular}
\label{tab:example1}
\end{table}

\begin{table}[ht]
\centering
\caption{Optimization schemes ablation study with a resnet-50 backbone on the kaokore status dataset.}\label{tab:table4}
\begin{tabular}{|l|l|l|l|l|}
\hline
Optimization method                  & Test accuracy & Validation accuracy & Test F1 & Validation F1 \\ \hline
First stage                          & 80.28         & 83.06               & 75.02   & 74.14         \\ \hline
Fully finetuned                      & 83.22         & 84.77               & 76.1    & 73.66         \\ \hline
First stage with Step LR             & 81.76         & 82.84               & 74.77   & 75.85         \\ \hline
Second stage fine-tuning with Step LR & \textbf{85.9}          & \textbf{87.34}               & \textbf{80.78}   & \textbf{81.18}         \\ \hline
\end{tabular}
\end{table}

We compare our method against other painting classification models in table \ref{table3}. Our optimal models utilizing ResNet and VGG architectures attain commendable results, comparable to the state-of-the-art LOOK model \cite{feng2021rethinking} and the five contrastive methods \cite{islam2021broad}, yet with significantly reduced computational demands. Our competitors adopt diverse strategies to evaluate the Kaokore dataset, employing an 8-way amalgamation of gender and status subclasses alongside validation metrics for benchmarking. Notably, these referenced works engage in full model fine-tuning, whereas our approach concentrates solely on fine-tuning the classifier head, showcasing the potency of our augmentation approach. While contrastive learning methods \cite{islam2021broad,feng2021rethinking} achieve superior test accuracy, it comes at the expense of protracted training and full-finetuning. In instances where they deviate from this approach, their performance lags behind our method. Additionally, these techniques exhibit considerable inefficiency in a few-shot setting, highlighting their data-intensiveness and computational burden. In contrast, our augmentation strategy seamlessly complements the state-of-the-art, operating as a preliminary step that could potentially enhance results in conjunction with existing methodologies.

Tweaking the proportions of rare and representative sample augmentations reveals distinct trends in model convergence and performance, as illustrated in Figure \ref{fig:trends}. Here, $p_1$ and $p_2$ represent the percentages of data from majority and minority classes, respectively, that are incorporated as extra training data to ensure stratified sampling. Capitalizing on the greater learning capacity of larger models, we focus our investigations on the ResNet-34 backbone within the spatial attention classifier. Notably, the training convergence of the model displays accelerated progress when utilizing fewer rare samples, a pattern consistently observed across various fixed $p_1$ values. Enhanced test accuracy is achieved through an increase in both rare and representative augmentations. In contrast, F1 scores demonstrate an advantageous response to a lower proportion of rare samples compared to representative ones. This trend opens up a balance between F1 scores for data bias mitigation and accuracy for class equilibrium. Moreover, it enables a trade-off between model convergence and potential overfitting, as the added rare samples contribute regularization.

Among the evaluated configurations (detailed in Table \ref{table3}), ResNet-50 stands out by achieving the most significant performance enhancement across all metrics when transitioning from no augmentation to the optimal augmentation mix ($p_1 = 0.3$ and $p_2 = 0.2$), thanks to its impressive 20 million trainable parameters. Furthermore, ResNet-50 showcases superior accuracy with diminished proportions of rare and representative samples compared to earlier setups.

Table \ref{prob_sweep} meticulously outlines the composite evaluation of model performance metrics, including accuracy, precision, and recall, across various combinations of rare and representative augmentations. The insights drawn from both Figure \ref{fig:trends} and Table \ref{prob_sweep} empower us to discern the judicious selection of rare proportions ($p_2$), which hinges upon the percentage of additional representative samples ($p_1$).

Moving to Table \ref{tab:table5}, we focus on evaluating model test accuracies using uniform hyperparameters from the best model variant. These hyperparameters include a learning rate of 0.00008, dropout probability of 0.23, weight decay of 0.0004, and L2 regularization. This evaluation centers on the 8-way combination \cite{islam2021broad} of status and gender subclasses within the Kaokore dataset. Importantly, each class in the supplementary dataset requires re-stylization prior to model training. The results depicted in the table reveal diminishing returns beyond the ResNet-101 architecture. Empirically, we discover that models exhibiting superior classification performance serve as better backbones for the classification task. For instance, VGG-19 outperforms VGG-16, and ResNet-101 surpasses other ResNet variations with fewer parameters. This trend suggests that pretrained models with higher capacities tend to yield better performance, with a discernible limit observed in very large backbones like ResNet-152. This is attributed to the insufficient capacity of the classifier head and spatial attention model when the backbone is frozen during the initial-stage fine-tuning.

Our experiments unveil the advantages of style transfer, particularly as model capacity increases, as demonstrated by the enhanced results in VGG-19 and ResNet-34 (Table \ref{tab:example1}). Notably, the control cases in the table do not utilize data augmentation. In this context, "data augmentation" refers to the optimal blend of stylized and original training data that yields the most favorable outcomes. Interestingly, we observe that larger models tend to overfit the dataset, making them more receptive to style transfer as a form of model regularization. Intriguingly, rare augmentations play a pivotal role for models with greater capacities, introducing more intra-class variations while mitigating overfitting. However, for smaller backbone models like VGG-16 and VGG-19, augmenting with representative samples is more effective. This strategy helps mitigate the drop in model performance due to excessive visual styles, as documented in \cite{zheng2019stada}. When examining tables \ref{prob_sweep} and \ref{tab:example1}, we observe that the data-augmented scenario demonstrates a narrower performance gap compared to the control case. This indicates enhanced class-specific performance, as evident from precision, recall, and F1 score metrics.

Table \ref{tab:table4} outlines an ablation study aimed at refining model performance through diverse fine-tuning strategies utilizing the ResNet-50 backbone. Optimal performance is achieved via a gradual unfreezing approach, encompassing both the backbone's two groups and the spatial attention mechanism in conjunction with the fully connected head. The model's progression involves initial freezing of the pretrained backbone, resulting in optimal head weights that yield heightened validation accuracy. Subsequently, in the second stage, the entire model is unfrozen and subjected to retraining, employing a diminished learning rate coupled with a multi-step learning rate scheduler that enacts decay every ten epochs. This approach facilitates comprehensive exploration at a broader global level within higher feature abstractions, while simultaneously permitting meticulous optimization space fine-tuning across the entire model distribution. Moreover, the hyperparameter search is bifurcated into two segments: an exponential scale coverage grid search followed by localized exploration centered around the most promising hyperparameter, determined using a Bayesian hyperparameter search methodology via Optuna, incorporating 100 exhaustive trials.

\begin{figure*}[!ht]
 \begin{subfigure}[b]{\textwidth}
   \includegraphics[width=\linewidth]{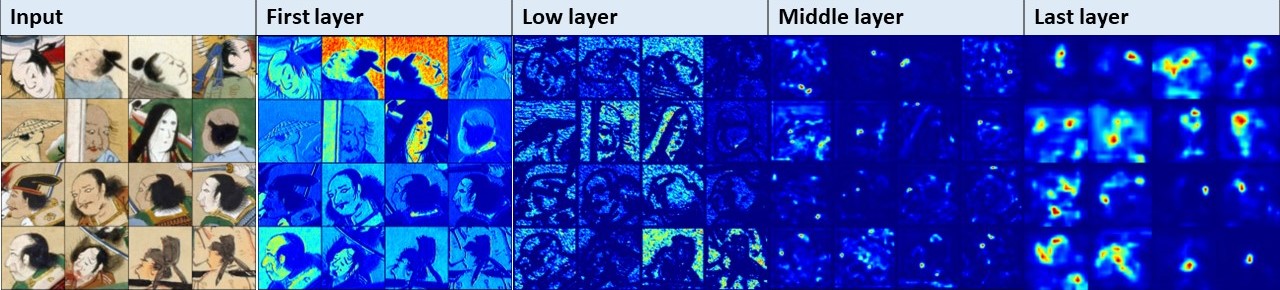}
  \caption{The attention map response without data augmentation for a random test batch.}
  \label{fig:amap_noaug}
 \end{subfigure}

 \begin{subfigure}[b]{\textwidth}
   \includegraphics[width=\linewidth]{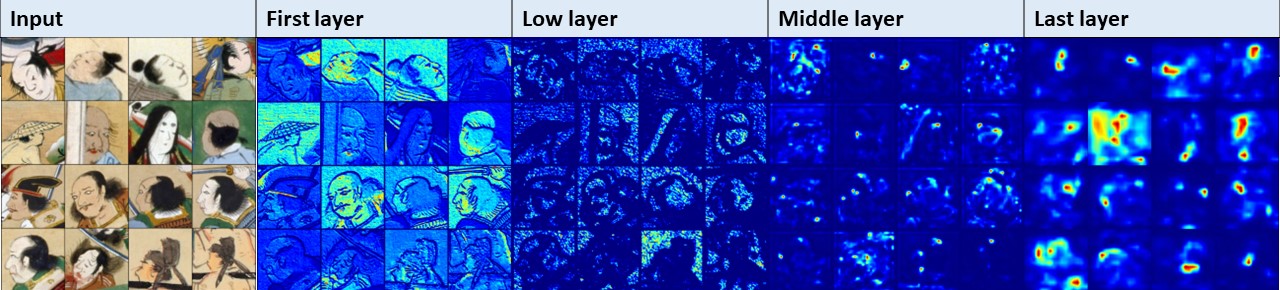}
  \caption{The attention map response with data augmentation for a random test batch.}
  \label{fig:amap_aug}
 \end{subfigure}

  \caption{Visualizing attention map responses of style transfer layers in a ResNet architecture \cite{vijendran23tackling}. The sequence, from left to right, includes input images and the outputs of spatial attention modules from the lowest, low, middle, and end classifier layers. Heat map response levels range from low to high, represented by shades progressing from dark blue to red.}
  \label{fig:amap}
 \end{figure*}

   \begin{figure*}[!ht]
  \centering
   \includegraphics[width=0.9\linewidth]{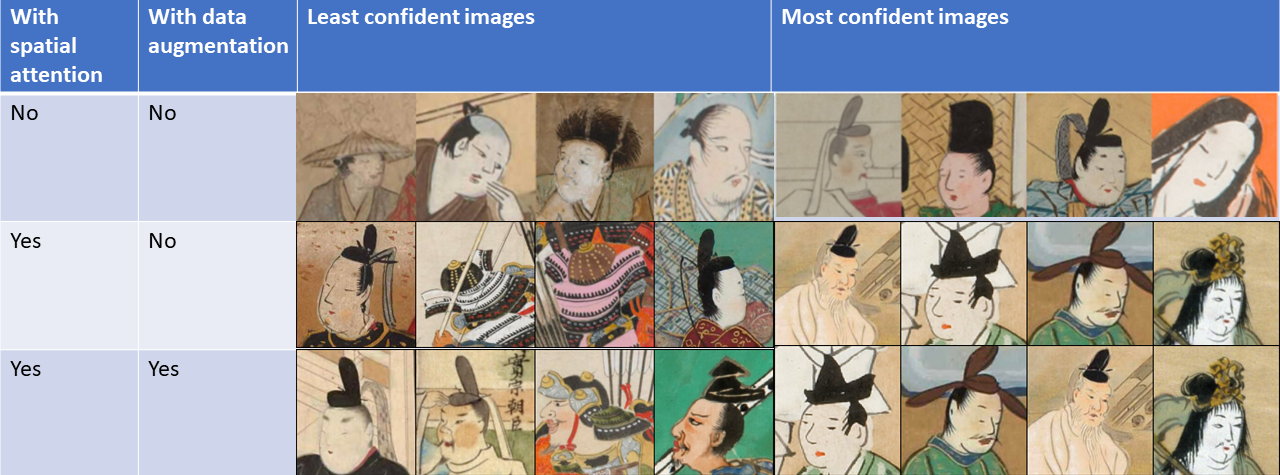}
   \caption{Visualizing least and most confident images from the validation subset of Kaokore dataset using various system configurations in the classifier with a VGG-16 backbone.}
  \label{fig:confident_imgages}
 \end{figure*}

\subsection{Qualitative Results}\label{subsec:qualres}

Spatial attention map visualizations, as shown in Figure \ref{fig:amap}, provide insights into the image regions that are crucial for the model's understanding. Comparing these visualizations with those in Figure \ref{fig:amap_noaug} (without data augmentation), it is evident that the model's attention becomes more focused and contrasts become clearer when data augmentation is applied. In the absence of augmentation, the model's attention is distributed over a broader area, with higher response levels in the lower layers of the model that are sensitive to texture, edges, and colors. In the Kaokore dataset, distinctive features for classifying different statuses include hairstyles, clothing, and facial parts, resulting in notable activation responses.

In Figure \ref{fig:amap_aug}, with data augmentation, we observe that texture details gain prominence over color information. Notably, regions containing faces and backgrounds retain high response levels, but as we move to later layers, the areas around hair and subjects become more significant. Overall, the response maps display enhanced contrast and higher response levels in the later layers when data augmentation is employed.

The set of least and most confident images, depicted in Figure \ref{fig:confident_imgages}, serves as a means to assess class imbalance. This selection is established by arranging model losses and visually presenting the top or bottom k images accordingly. The arrangement progresses from the least to the most confident images, based on their corresponding losses. The choice of the VGG-16 model was informed by its compatibility with style transfer as a backbone. For the version with style augmentation, the least confident images predominantly include examples from the noble class due to the sampling bias towards the noble class within the test set. Conversely, in the absence of data augmentation and spatial attention, the least confident images are from the commoner class, revealing class imbalance despite its smaller test sample size. The other two configurations share a similar set of images, albeit with varying degrees of contrast and detail in the most confident images. The introduction of spatial attention alone results in the least confident images containing complex backgrounds and subjects with obscured faces. On the other hand, style transfer-based augmentations address this particular weakness, but they may not effectively handle highly complex backgrounds or rare poses. By providing style variations for each sample, the model can focus on texture invariance while potentially overlooking intricate image details.

\subsection{Implementation Details}\label{subsec:implementationdets}

In the pre-training phase, the style transfer model is trained on pairs of style and content images drawn uniformly from the training dataset. This process involves 20,000 iterations per class. Subsequently, the learned decoder is utilized for inference, generating stylized counterparts through similar sampling of class-specific style and content pairs. This augmentation dataset maintains an in-class distribution identical to the training dataset, ensuring an equal number of samples for each class. Our experimental setup adheres to the parameters of the AdaIN style transfer network \cite{huang2017arbitrary}.

For the model training, a batch size of 64 is employed over 20 epochs, utilizing an Adam optimizer. Dropout and a focal loss are incorporated, with the gamma and alpha values set to 2 following recommended guidelines \cite{Lin_2017_ICCV}. To expedite data processing, 8 CPU workers are utilized.

Leveraging L2 and particularly L1 regularization in conjunction with the focal loss helps the model to accentuate specific parts of the features. This is especially crucial since style transfer can enhance finer details from features across different levels of granularity that might otherwise be diminished by convolution and pooling operations. The inclusion of dropout further amplifies the degree of model regularization. Throughout both the pre-training phase to generate augmented data counterparts and the inference during classifier training, a single NVIDIA A100 GPU is utilized.

In our classifier, we integrate pre-trained models including ResNet (from ResNet-34 to ResNet-152) and VGG (VGG-16 and VGG-19) variants \cite{canziani2016analysis,8745417}. These selections enable a comparative analysis against the Kaokore dataset benchmarks \cite{tian2020kaokore}. The introduction of architecture variants permits the examination of augmentation impacts on model capacity. Notably, their weights remain frozen at the classifier level to underscore the influence of data augmentation rather than the inherent model architecture. The pre-trained models retain their fully connected layers but retain the last layer as a global average pooling layer. This adaptation enhances the model's robustness to images of varying sizes and enhances its role as a feature extractor.

%% file: sections/conclusions-FL.tex
\section{Conclusions}
\label{sec:conclusion}


Our innovative approach harnesses style transfer to generate classifier-specific stylized images, resulting in superior outcomes compared to models trained on unaugmented datasets. Through strategic manipulation of augmented samples' proportions in minority and majority classes, we achieve a delicate balance between model convergence speed and performance enhancement. Our system capitalizes on a two-stage process, wherein a stylized training dataset feeds into a spatial attention-equipped classifier. This foundation is further optimized via two rounds of hyperparameter search and model fine-tuning. We expertly navigate the trade-off between accuracy and convergence while considering recall, precision, and F1 score by modulating the extra data ratio for both minority and majority classes. Introducing 20-60\% rare augmentations to minority classes bolsters recall, precision, and F1 scores, while augmenting representative samples by 50-90\% results in across-the-board metric improvements, particularly accentuating accuracy and model convergence. Our approach's efficacy is affirmed through qualitative examinations of class imbalance and backbone interpretability across various layers, followed by quantitative analyses spotlighting weak supervision cues from spatial attention modules and mitigated data bias through style transfer augmentations. The zenith of our achievement lies in demonstrating the performance surge facilitated by a two-stage hyperparameter search, complemented by fine-tuning procedures incorporating gradual unfreezing and initial learning rate reduction.



Our novel method attains superior performance by harnessing automated style image generation for style transfer. The process involves random sampling of style and content images within each class, creating an augmented dataset that may not necessarily provide challenging examples for optimal focal loss utilization during training. Looking ahead, we envision the integration of stylized sample generation into our two-stage optimization approach. This strategic enhancement would furnish the model with harder samples at distinct stages of fine-tuning, further enhancing its capabilities.


Additionally, there is potential for further exploration in incorporating spatial attention with geometric priors, like landmarks or segmentation masks, or by utilizing the style transfer latent code. This approach could lead to feature correspondences that are either semantically or functionally significant. Masks representing content or style could guide attention towards specific features within images. Leveraging the classifier's ability to learn texture or shape could also extend to image retrieval applications.


Finally, we plan to investigate the model's generalization capabilities on diverse painting datasets with distinct styles, e.g., WikiArt \cite{saleh2015large} or the Ukiyo-e dataset \cite{pinkney2020ukiyoe}. The WikiArt dataset comprises artworks spanning various genres and styles, including a subset of Japanese art akin to the present study. This dataset would enable us to compare our approach across different artistic traditions. Also, the Ukiyo-e faces dataset has the potential to extend the scope of our work to other tasks, such as facial landmark detection, where multiple subjects and varying painting compositions are present.